\documentclass{article} 
\usepackage{iclr2025_conference,times}

\usepackage{wrapfig}
\usepackage{lipsum} 

\usepackage{amsmath,amsfonts,bm}
\usepackage{multicol}

\def\eqref#1{equation~\ref{#1}}

\def\1{\bm{1}}

\DeclareMathAlphabet{\mathsfit}{\encodingdefault}{\sfdefault}{m}{sl}
\SetMathAlphabet{\mathsfit}{bold}{\encodingdefault}{\sfdefault}{bx}{n}

\usepackage{hyperref}
\usepackage{url}
\usepackage{multicol}
\usepackage{aliascnt}
\usepackage{adjustbox}
\usepackage{booktabs}
\usepackage{multirow} 
\definecolor{uclablue}{rgb}{0.15, 0.45, 0.68}
\hypersetup{
    breaklinks,
    citecolor=uclablue,
    colorlinks=true,
}
\title{A Comparative Study on Reasoning Patterns of OpenAI's o1 Model}

\newcommand*\samethanks[1][\value{footnote}]{\footnotemark[#1]}

\author{Siwei Wu$^{1,2,3}$\thanks{Equal Contribution.}, Zhongyuan Peng$^{7}$\samethanks[1], Xinrun Du$^{1}$\samethanks[1], Tuney Zheng$^{1,3}$\samethanks[1], Minghao Liu$^{4}$, Jialong Wu$^{1}$,\\ 
\bf Jiachen Ma$^{6}$, Yizhi Li$^{1,2}$ Jian Yang$^{1}$, Wangchunshu Zhou$^{1,3}$, Qunshu Lin$^{5}$,\\
\bf Junbo Zhao$^{6}$, Zhaoxiang Zhang$^{7}$, Wenhao Huang$^{1}$, Ge Zhang$^{1,3}$\thanks{Corresponding Authors.}, Chenghua Lin$^{1,2}$\samethanks[2], J.H. Liu$^{1}$\samethanks[2]\\
 $^1$M-A-P, $^2$University of Manchester, $^3$OpenO1 Team, $^4$ 2077AI, $^5$ Abaka AI, \\
 $^6$ Zhejiang University, $^7$ University of Chinese Academy of Sciences
}

\begin{document}
\maketitle

\begin{abstract}
Enabling Large Language Models (LLMs) to handle a wider range of complex tasks (e.g., coding, math) has drawn great attention from many researchers. 
As LLMs continue to evolve, increasing the number of model parameters yields diminishing performance improvements and heavy computational costs.
Recently,
OpenAI's o1 model has shown that inference strategies (i.e., Test-time Compute methods) can also significantly enhance the reasoning capabilities of LLMs. However, the mechanisms behind these methods are still unexplored.
In our work, to investigate the reasoning patterns of o1, we compare o1 with existing Test-time Compute methods (BoN, Step-wise BoN, Agent Workflow, and Self-Refine) by using OpenAI's GPT-4o as a backbone on general reasoning benchmarks in three domains (i.e., math, code and commonsense reasoning).
Specifically,
first,
our experiments show that the o1 model has achieved the best performance on most datasets.
Second,
as for the methods of searching diverse responses (e.g., BoN), we find the reward models' capability and the search space both limit the upper boundary of these methods.
Third,
as for the methods that break the problem into many sub-problems, the Agent Workflow has achieved better performance than Step-wise BoN due to the domain-specific system prompt for planning better reasoning processes.
Fourth,
we summarize six reasoning patterns of o1, and provide a detailed analysis across different reasoning benchmarks.
Finally, code and dataset are released in \url{https://github.com/Open-Source-O1/o1_Reasoning_Patterns_Study}.

\end{abstract}

\section{Introduction}

Large Language Models (LLMs) have achieved great success in various tasks (e.g., Commonsense Reasoning~\citep{yang2018hotpotqa}, Coding~\citep{jain2024livecodebench,chai2024mceval}, Math~\citep{satpute2024can,chai2024xcot}, and Dialogue~\citep{young2023investigating}). To further improve their performance, researchers have continuously increased the number of model parameters and expanded the training data. However, this method of scaling up the model parameters is reaching a bottleneck, and the efficiency of performance improvement is becoming progressively limited.

Recently, Test-time Compute methods, such as Best-of-N (BoN) and Self-Refine~\citep{madaan2024self}, have been proposed to enhance model performance during the inference phase and have shown to be more efficient than simply increasing model parameters. However, there is a lack of research comparing the effectiveness of different Test-time Compute methods across various tasks, which would provide valuable guidance for researchers developing new models. 
Besides,
understanding the inference mechanism of the o1 model is very important to help researchers enhance the capabilities of LLMs.

To address the aforementioned issues, we compare OpenAI's o1 model with various Test-time Compute methods, using GPT-4o as the backbone. 
According to the OpenAI o1 report~\footnote{\url{https://openai.com/index/introducing-openai-o1-preview/}}, the model demonstrates exceptional improvements in areas such as mathematics and coding. 
Therefore, we select four benchmarks—HotpotQA~\citep{yang2018hotpotqa}, Collie~\citep{yao2023collie}, USACO~\citep{shi2024can}, and AIME~\footnote{\url{https://huggingface.co/datasets/AI-MO/aimo-validation-aime}}—to encompass three key reasoning domains.
For certain benchmarks (i.e., HotpotQA and Collie) that are not challenging for current LLMs,
we follow the LIME~\citep{zhu2024lime} and implement a voting method using four selected models (i.e., Qwen~\citep{bai2023qwen,yang2024qwen2}, Yi~\citep{ai2024yi}, Llama3~\citep{dubey2024llama}, and Claude~\footnote{\url{https://claude.ai/}}) to filter out samples that cannot be correctly answered by more than two of the LLMs.
Then we select four Test-time Compute methods (including Best-of-N (BoN), Step-wise BoN, Agent Workflow, and Self-Refine) as baselines which use the GPT-4o as the backbone.
As for BoN and Step-wise BoN, we use GPT-4o as the reward model to select the most suitable responses for a given sample.
We directly use the code from the GitHub of the Self-Refine~\citep{madaan2024self}.
As for the Agent Workflow, we utilize the state-of-the-art agent framework~\citep{zhou2024symbolic} on the HotpotQA and Collie, and we use the GPTs~\footnote{\url{https://openai.com/index/introducing-gpts/}} for USACO and AIME.

We have conducted comprehensive experiments on our filtered benchmarks, and we have the following insightful findings:
\begin{itemize}
    \item The OpenAI's o1 model achieves the best results across almost all benchmarks and demonstrates significant improvements in coding and math tasks using the CoT-based approach. 
    \item The domain-specific system prompt is crucial for Step-wise methods. Specifically, the Agent Workflow method greatly enhances the model's performance and it is relatively close to the o1's performance, while the impact of Step-wise BoN on the model's capabilities is mainly evident in the HotpotQA task.
    Besides, we assume that the Agent Workflow with a series of domain-specific system prompts can not only reduce unnecessary reasoning steps but also carefully align with the reasoning problems.
    
    
    \item We summarize 6 types of o1 reasoning patterns (i.e., \textbf{Systematic Analysis (SA)}, \textbf{Method Reuse (MR)}, \textbf{Divide and Conquer (DC)}, \textbf{Self-Refinement (SR)}, \textbf{Context Identification (CI)}, and \textbf{Emphasizing Constraints (EC)}) across four benchmarks, and we observe that the most commonly used reasoning patterns in o1 are DC and SR, which might be the key to o1's success. 
    Moreover, the reasoning patterns vary across different tasks.
    Specifically, for commonsense reasoning tasks, o1 tends to use CI and EC. In contrast, in math and coding tasks, o1 mainly relies on MR and DC. 
    \item We also explore the number of reasoning tokens of o1  across different tasks, and observe that the number of reasoning tokens varies a lot across different tasks.
\end{itemize}



\section{Related Work}

\subsection{Large Language Models}
With the emergence of Transformers~\citep{vaswani2017attention} and the scaling laws~\citep{henighan2020scaling}, the researchers try to scale up the parameters of the generative language model. 
As a result, OpenAI's GPT series models~\citep{radford2018improving,radford2019language,brown2020language,achiam2023gpt} have achieved remarkable success in the NLP field.
Inspired by the scaling law, the rapid development of open-source models has also been achieved through scaling up the size of parameters and collecting huge data for pre-training, such as Qwen~\citep{bai2023qwen,yang2024qwen2}, Yi~\citep{ai2024yi}, Llama~\citep{touvron2023llama,dubey2024llama}, and Deepseek~\citep{bi2024deepseek}.
Apart from these, current researchers are meeting the demands of training LLMs by collecting higher-quality instruction data and pre-training data.
Moreover, improving the quality of the collected data has also gained significant attention in developing LLMs. However, the approach of enhancing model performance by increasing model parameters and collecting more data is facing a bottleneck~\citep{snell2024scaling}.

\subsection{Test Time Compute Methods}
\citet{snell2024scaling} propose that scaling LLMs Test-time Compute optimally can be more effective than scaling model parameters. 
Besides,
there are some methods designed for adapting Test-time Compute to LLMs' reasoning.
OpenAI's o1 model~\footnote{\url{https://openai.com/o1/}} is designed to spend more time reasoning before they respond for the sake of obtaining better performance. 
\citet{wang2023hypothesis} conduct a hierarchical hypothesis search to enable inductive reasoning capabilities.
Besides,
A number of related works have been proposed to augment LLMs with tools with Test-time Compute, which can greatly improve their performance on downstream tasks~\citep{gao2023pal,qin2023toolllm,qu2024tool}.
Moreover, several works have been proposed to learn thought tokens in an unsupervised manner~\citep{goyal2023think,zelikman2024quiet}, which enable models to more effectively utilize the Test-time compute with sampling longer sequences.
In this work, we explore the performance of OpenAI's o1 model on several common NLP reasoning tasks and investigate the reasoning patterns when compared to some classical Test-time Compute methods.

\section{Experimental Setup}
In order to comprehensively evaluate the capability of OpenAI's o1 model, we select and filter 4 benchmarks covering 3 domains (i.e. Commonsense Reasoning, Math, and Code). Then we provide the results of o1, GPT-4o, and some traditional Test-time Compute methods.

\subsection{Benchmarks}

\paragraph{Commonsense Reasoning.}
We select \textbf{HotpotQA}~\citep{yang2018hotpotqa} and \textbf{Collie}~\citep{yao2023collie} to evaluate the commonsense reasoning ability of LLMs. The HotpotQA mainly focuses on commonsense reasoning, which requires LLMs to use multiple supporting documents to answer.
Collie needs LLMs to generate text allowing the specification of rich, compositional constraints with diverse generation levels.
Due to the excellent open-ended response generation capabilities of GPT-4o and o1, these models demonstrate relatively strong performance on certain benchmarks, particularly in commonsense reasoning. According to LIME~\citep{zhu2024lime}, we design a \textbf{data filtering} module to show the performance differences among different models. This module involves using four different LLMs (i.e., Llama3-72B~\citep{dubey2024llama}, Qwen-72B~\citep{bai2023qwen}, Claude to answer each sample in those benchmarks and subsequently filtering out samples that more than two models can answer correctly.

\paragraph{Code.}
We are using the bronze level of the \textbf{USACO}~\citep{shi2024can} competition to test the coding skills of LLMs. The USACO focuses on algorithmic and problem-solving skills. We employ LLMs like Llama3-72B, Qwen-72B, and Claude to solve these problems, selecting only those that prove challenging across multiple models to ensure a rigorous assessment of their coding abilities.

\paragraph{Math.}
We directly use the AIME~\footnote{\url{https://huggingface.co/datasets/AI-MO/aimo-validation-aime}} benchmark to evaluate the model's math ability, which contains 90 problems from AIME 22, AIME 23, and AIME 24, and have been extracted directly from the AOPS wiki page.

\subsection{Baseline methods}

We select two powerful closed-source LLMs for evaluation.
\paragraph{o1 model.} It is designed to spend more time reasoning before they respond, which can reason through complex tasks and solve harder problems than previous models in science, coding, and math.
\paragraph{GPT-4o.} It is a multimodal model that integrates text, vision, and audio processing capabilities into a single and unified neural network. 

As for Test-time Compute methods, we select four methods based on GPT-4o.

\paragraph{Best-of-N (BoN).} It makes LLMs generate multiple $N$ outputs for a given input, and the most suitable response is selected as the output.

\paragraph{Step-wise BoN.} It enables LLMs to analyze a problem and break it down into several sub-problems. For each step, the model generates $N$ responses based on the previous sub-problems and answers, and then we use a reward model to select the best response. This process continues iteratively until the final answer to the original problem is obtained.

\paragraph{Self-Refine.} It improves initial outputs from LLMs through iterative feedback and refinement~\citep{madaan2024self}.

\paragraph{Agent Workflow.} LLM agents break down complex tasks into smaller sub-tasks, plan their execution through a structured workflow, and utilize various tools to achieve their goals. 
For the commonsense reasoning datasets, we leverage the existing state-of-the-art agent framework~\citep{zhou2023agents,zhou2024symbolic} for evaluation.
For the code and math datasets, we select the top-picked agents from GPTs~\footnote{\url{https://openai.com/index/introducing-gpts/}}, specifically \textit{code copilot} and \textit{math solver}, respectively.

\subsection{Metrics}
As for HotpotQA and AIME, we design a rule to determine whether the model-generated response contains the correct answer and use the accuracy of the model's responses as the final score.
Regarding Collie, we directly determine whether the model-generated response is correct.
As for coding tasks (i.e., USACO), we manually run the LLMs-generated code on the test examples, and regard the code passing the test cases as right.

\section{Results}
\begin{table*}[hbt!]
\begin{center} 
\footnotesize
\resizebox{0.99\columnwidth}{!}{
    \begin{tabular}{l|l|l|l|cc|c|c }
\toprule
\multicolumn{1}{l|}{ }                                                                     & \multicolumn{1}{l|}{ } &  \multicolumn{1}{l|}{ } &\multicolumn{1}{c|}{ }                                                                              & \multicolumn{2}{c|}{\textbf{Commonsense Reasoning}}                                                                           & \multicolumn{1}{c|}{\textbf{Code}}     & \multicolumn{1}{c}{\textbf{Math}}                                                                        \\
\multirow{-2}{*}{\textbf{Setting}}  & \multirow{-2}{*}{\textbf{Baselines}}                                                                      & \multirow{-2}{*}{\textbf{$N$}}  &  \multirow{-2}{*}{\textbf{Overall}}  &
HotpotQA&	Collie  &	USACO &	AIME \\

\midrule\midrule
\multirow{-0}{*}{\textbf{Direct}}  & o1-preview  & - & 34.32
 & 14.59	 &  34.07	 &  \textbf{44.60}	 &   44.00	 	
 \\ 
 & o1-mini  & - &  \textbf{35.77} 	 &  15.32	 &   \textbf{53.53}	 &   12.23  & 	\textbf{62.00 }	
 \\ 
  & GPT4o & - & 18.44  & 13.14 	& 43.36 	&  \ \ 5.04 	& 12.22

 \\
 \midrule\midrule
 
 & BoN  & 4  &  17.65  & 	13.50 	 & 39.82	 &   \ \ 5.04  & 	12.22

  \\

& BoN  & 8 &	19.04 & \textbf{16.42} &	38.50 &   \ \ 7.91	&13.33 	
\\

& Step-wise BoN & 1 & \ \ 6.09 &	13.50 &	\ \ 5.31 &	   \ \ 0.00	& \ \ 5.56	
\\
 
& Step-wise BoN  & 4 &  \ \ 9.79	& 15.69	&  19.55 	&   	\ \ 0.00 	& \ \ 7.78
 \\

& Self-Refine  & 3 &  \ \ 5.62	& 13.25	& \ \ 0.00 	&  	\ \ 0.00	& \ \ 9.23
 \\

\multirow{-5.5}{*}{\textbf{Test-Time }}  & Agent Workflow & -  & 24.70 & 14.96 	& 46.07	&   22.22 & 	15.56  
 \\

\bottomrule
\end{tabular}
}
\end{center}
\caption{
The results of OpenAI's o1 model, GPT4o, and some Test-time Compute methods on our selected four benchmarks (i.e., HotpotQA, Collie, USACO, AIME). The `-' in the table represents that the method does not search the multiple responses for generation. 
\textbf{Direct} refers to having the LLMs generate a response directly from the input text, while \textbf{Test-Time} refers to using the Test-time Compute method based on GPT-4o.
}
\label{tab:Main Result}
\end{table*}

\subsection{Overall Analysis}

We conduct various experiments to evaluate the performance of o1 and the Test-time Compute methods.
As shown in Table~\ref{tab:Main Result}, the OpenAI's o1 model achieves the best performance on most benchmarks compared to previous Test-time Compute methods and GPT-4o, particularly in Math and Code tasks.
Among those benchmarks, o1's improvement in mathematical and coding tasks is particularly notable compared to other methods, which shows that this thinking-before-reasoning approach is more suitable for complex multi-step reasoning in mathematical and coding tasks.
Specifically, the o1-mini surpasses the o1-preview on some tasks, it shows that the reasoning process of o1 does not always lead to better improvements.

\paragraph{The performance improvement from Self-Refine is not significant.}
On most tasks, Self-Refine shows only a slight improvement compared to GPT-4, and its performance even declines on Collie.
For this phenomenon,
we assume that LLMs may generate responses that slightly deviate from the required format during the refinement iterations of Self-Refine.

\paragraph{BoN achieves relatively good results on HotpotQA.} It demonstrates the necessity of searching for more possible responses during the inference stage by scaling time.
However, the performance of BoN on Collie has declined compared to the original GPT-4o. 
Besides,
when $N$ increases, there is a slight degradation in performance.
We believe this is due to Collie's strict format requirements, which limit the effectiveness of diverse outputs from LLMs. 

\paragraph{The Step-wise BoN is limited by the complex tasks.}
As for Step-wise BoN, it achieves an excellent result on HotpotQA, which does not have a restriction on output text. 
However, its performance drops significantly on other complex benchmarks that make Step-wise BoN generate numerous intermediate steps and cannot follow the original question.

\paragraph{Agent Workflow achieves a significant improvement in performance on all benchmarks.} 
The Agent Workflow uses a similar idea to the step-wise BoN that breaks down complex tasks into smaller subtasks, but it designs a series of domain-specific system prompts,
which reduces unnecessary long-context reasoning processes. 
However, there is still a gap between the Agent Workflow and the o1 model, which may be because Agent Workflow explores a less diverse space of responses.


\subsection{Analysis of the reasoning pattern of o1}
As shown in Table~\ref{tab:Main Result}, although o1 is generally much better than other models, some Test-time Compute methods can still achieve relatively close results to o1 in certain specific tasks.
To this end, we analyze the reasoning patterns of  o1 across various tasks and summarize the reasoning patterns across different benchmarks as follows:

\begin{itemize}
    \item \textbf{Systematic Analysis (SA).} Starting from the overall structure of the problem, o1 first analyzes the inputs and outputs, as well as the constraints, and then decides on the choice of algorithm and the use of data structures.
    \item \textbf{{Method Reuse} (MR).} For some problems that can be transformed into classic problems (such as the shortest path or knapsack problem), o1 can quickly reuse existing methods to solve them.

    \item \textbf{Divide and Conquer (DC).} It breaks down a complex problem into subproblems and constructs the overall solution by solving the subproblems.

    \item \textbf{Self-Refinement (SR).} o1 assesses its reasoning process during inference to determine if there are any issues and correct any errors.

    \item \textbf{Context Identification (CI).} For some datasets requiring additional information input (e.g., HotpotQA),
    o1 first summarizes different aspects of the context related to the query, and then gives the response for the corresponding query.
    
    \item \textbf{Emphasizing Constraints (EC).} For some datasets with constraints on the generated text (e.g., Collie), o1 usually emphasizes the corresponding constraints during the reasoning process.
    
\end{itemize}




\begin{figure}[t]
    \centering
    \includegraphics[width=0.75\linewidth]{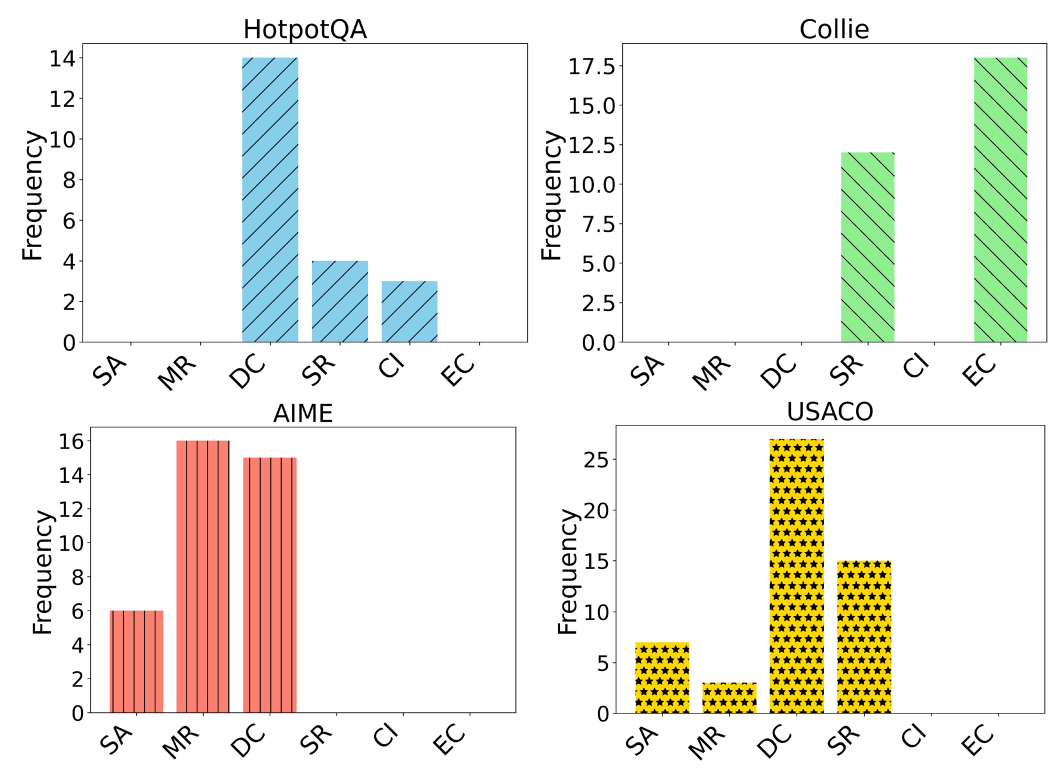}
    \caption{The statistics of different reasoning patterns on different benchmarks. }
    \label{fig:Thinkling_pattern_on_benchmarks}
\end{figure}

\begin{wrapfigure}{r}{0.45\textwidth}
    \centering
    \includegraphics[width=0.99\linewidth]{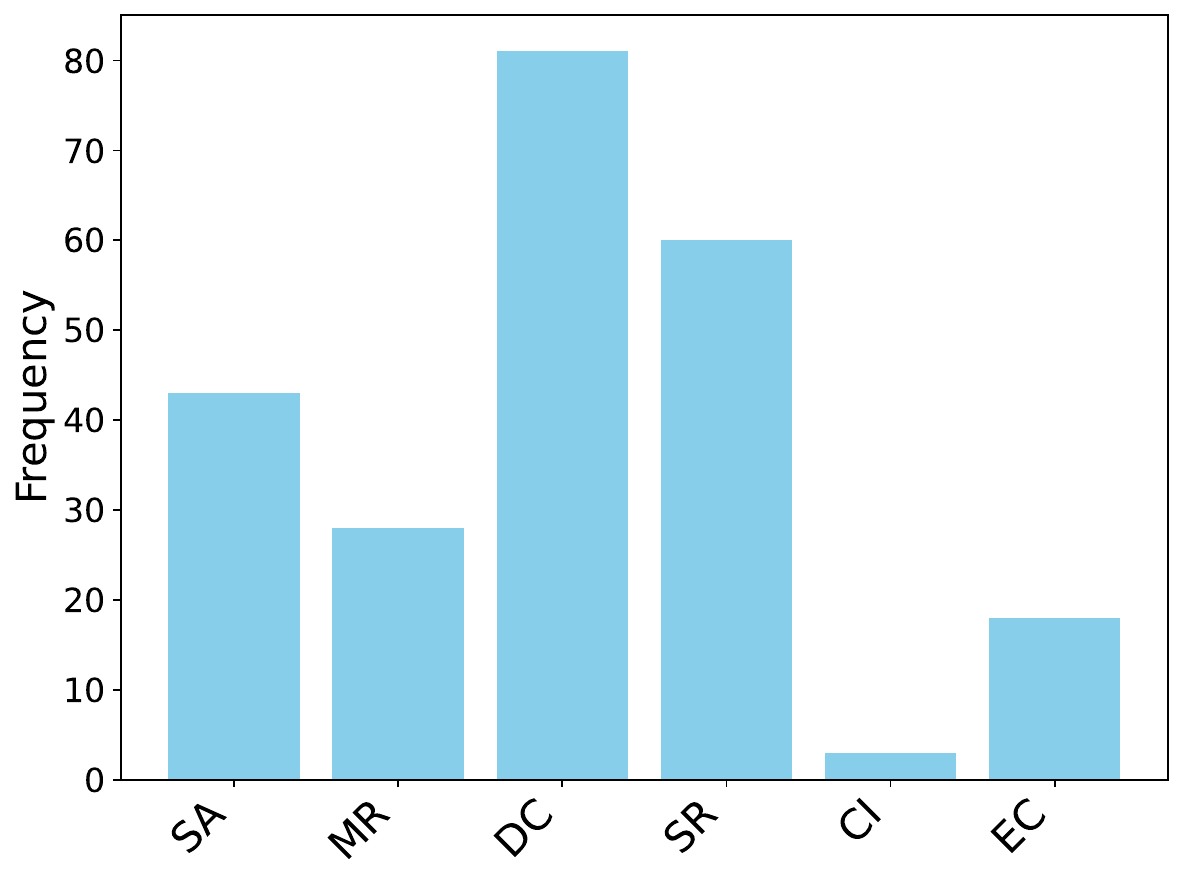}
    \caption{The statistics of reasoning patterns. }
    \label{fig:thinking pattern}
\end{wrapfigure}

We randomly selected 20 to 30 samples of each benchmark to count the number of different reasoning patterns. 
As shown in Fig.~\ref{fig:thinking pattern}, the performance of o1 is primarily influenced by three reasoning patterns: DC, SR, and SA. Among these, SA and DC appear most frequently, suggesting that the combination of SR and DC plays a crucial role in enhancing the performance of o1.

We also show the statistics of the reasoning patterns on different benchmarks in Fig.~\ref{fig:Thinkling_pattern_on_benchmarks},
where different tasks require different reasoning patterns.
Specifically,
in commonsense reasoning tasks, o1 tends to use task-specific global analysis methods (such as CI and EC) and DC. In math and coding tasks, o1 mainly relies on DC and MR.
For both Collie and AIME, o1 follows a relatively shorter reasoning process, which we find is also linked to its reasoning patterns. Specifically, o1 often employs the MR approach, where it directly applies well-known classic solutions to solve mathematical problems without the need for multi-step reasoning. 
In the case of Collie, o1 tends to use the EC reasoning pattern. 
This allows the model to place greater emphasis on Collie's output format requirements, preventing the generation of an excessively long reasoning process that would result in outputs not meeting the format requirement.

\subsection{Long Context Inference Limits Step-Wise BoN}
Apart from generating multiple responses in breadth, the Step-wise strategy is also important for scaling inference time. 
Specifically, the Step-wise methods often produce many intermediate steps, and excessively long context information can prevent the model from following the original input text to generate the correct response.
As shown in Table~\ref{table:reasoning token}, we provide the average number of tokens in the intermediate steps of Step-wise BoN inference across different tasks. The average number of reasoning tokens in almost all tasks exceeded 200, which also confirms that Step-wise BoN requires the model to have strong long-context following capabilities.
The Step-wise BoN performs relatively worse on tasks like Collie and AIME, where the output text format and reasoning process are highly complex (for instance, Step-wise BoN achieves less than 12\% accuracy on Collie, and its performance on AIME is only half that of other methods). However, for tasks (e.g., HotpotQA) that do not require stringent output formatting or intricate reasoning, both BoN and Step-wise BoN significantly enhance the model's results (when $N=4$, Step-wise BoN outperforms GPT-4o by 2.55\% and BoN surpasses GPT-4o by 0.36\% on HotpotQA).

\begin{wraptable}{r}{0.51\textwidth}
\centering
\small
\begin{tabular}{cc|c|c}
\toprule
 \multicolumn{2}{c|}{\textbf{Commonsense Reasoning}} & \multicolumn{1}{c|}{\textbf{Coding}} & \multicolumn{1}{c}{\textbf{Math}}  \\
 HotpotQA & Collie & USACO & AIME \\
\midrule
\midrule
273.59 & 450.31 & 439.90 & 262.51 \\
\bottomrule
\end{tabular}
\caption{The average reasoning token length of Step-wise BoN ($N=4$).}
\label{table:reasoning token}
\end{wraptable}

\subsection{The Number of Reasoning Tokens Across Different Tasks for o1}

\begin{figure}[t]
    \centering
    \includegraphics[width=0.6\linewidth]{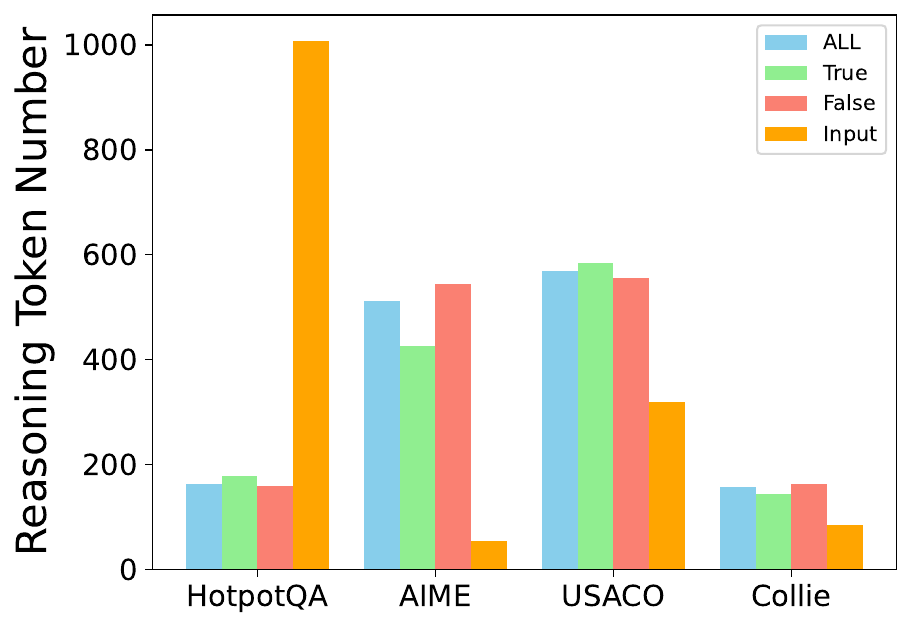}
    \caption{
   The statistics of the number of o1's reasoning tokens on different tasks. `ALL' represents the average length of reasoning tokens for all samples, while `True' and `False' show the averages for correctly and incorrectly answered samples, respectively. `Input' refers to the average length of the input prompt.
    }
    \label{fig:reasoning token}
\end{figure}

To investigate whether the number of reasoning tokens is related to o1's ability, we developed a rule to extract o1's reasoning tokens and computed their count across different tasks. Additionally, we calculated the average number of tokens for both correct and incorrect samples.
Furthermore, to explore the relation between input prompt length and reasoning tokens length, we also calculate the average input length.
As shown in Fig.~\ref{fig:reasoning token}, 
we observe that the number of reasoning tokens for correct and incorrect samples is similar for the same task, and there is no clear correlation between the input prompt length and the length of the reasoning tokens.
Instead, there is a significant difference in reasoning tokens across different tasks.
Specifically, for commonsense reasoning tasks (i.e., HotpotQA and Collie), the o1's reasoning token length is relatively short. However, for more difficult tasks like Code (i.e., USACO) and Math (i.e., AIME), the model often requires a longer reasoning process to obtain the correct answer.

\subsection{The Reward Model Limits Abilities of Searching Methods}

\begin{figure}[htbp]
    \centering
    \begin{minipage}[b]{0.45\textwidth}
        \centering
        \includegraphics[width=\textwidth]{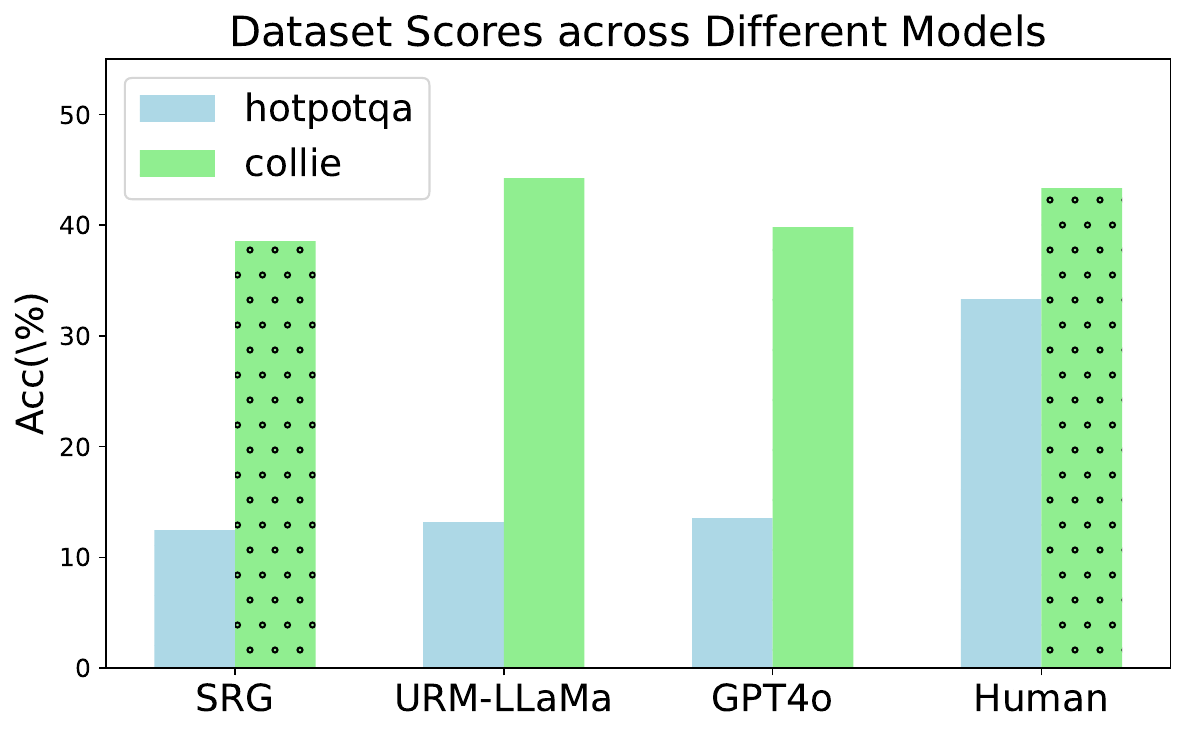}
        \caption{The results of BoN( GPT-4o) using different reward models under $N = 4$ setting. The SRG represents the Skywork-Reward-Gemma-2-27B, the URM-LLaMa refers to the URM-LLaMa-3.1-8B.}
        \label{fig:reward_model}
    \end{minipage}
    \hspace{0.05\textwidth} 
    \begin{minipage}[b]{0.4\textwidth}
        \centering
        \includegraphics[width=\textwidth]{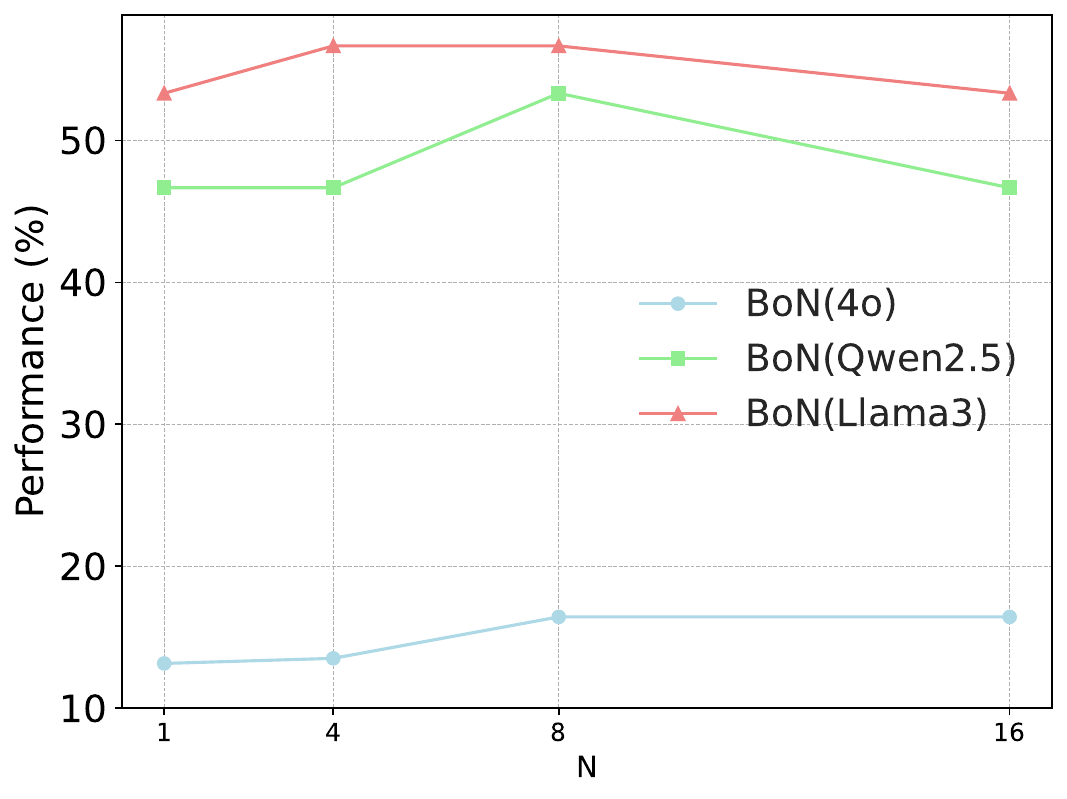}
        \caption{The results of BoN under different search spaces (i.e. the $N$ ranging from 1 to 16) on HotpotQA.}
        \label{fig:BoN}
    \end{minipage}
\end{figure}

As for the BoN series methods, they need to use a reward model to choose the most suitable responses among all the generated responses. 
Especially for the Step-wise method, an error in any intermediate step can lead to error accumulation, which significantly affects the final output of the model.
Therefore, for BoN, we conduct experiments by using different reward models (e.g. Skywork-Reward-Gemma-2-27B~\citep{skyworkreward2024} and URM-LLaMa-3.1-8B~\citep{lou2024uncertaintyawarerewardmodelteaching}) from the Leaderboard of RewardBench~\citep{lambert2024rewardbench}.
We also use the GPT-4o as the reward model to choose the most suitable response.
Moreover,
to demonstrate the reward models' limitation on the searching ability of LLMs, we also use the Human as the reward model to judge the most suitable generated response of BoN.
As shown in Fig.~\ref{fig:reward_model}, we provide the results of BoN (GPT-4o) using different reward models.
As for HotpotQA, those methods have relatively worse performance (i.e., their accuracies are under 15\%), while the human-based reward model could help to improve the LLMs' ACC to 33\%.
As for the Collie, the performances of using other reward models are close to human results, which show the reward models' ability to determine the upper boundary of those methods.
Besides, although the BoN could have comparable results on HotpotQA compared with o1, it can be significantly improved by using a more powerful reward model, demonstrating that the reward model is crucial for the search methods.

\begin{wrapfigure}{r}{0.45\textwidth}
    \centering
    \includegraphics[width=0.99\linewidth]{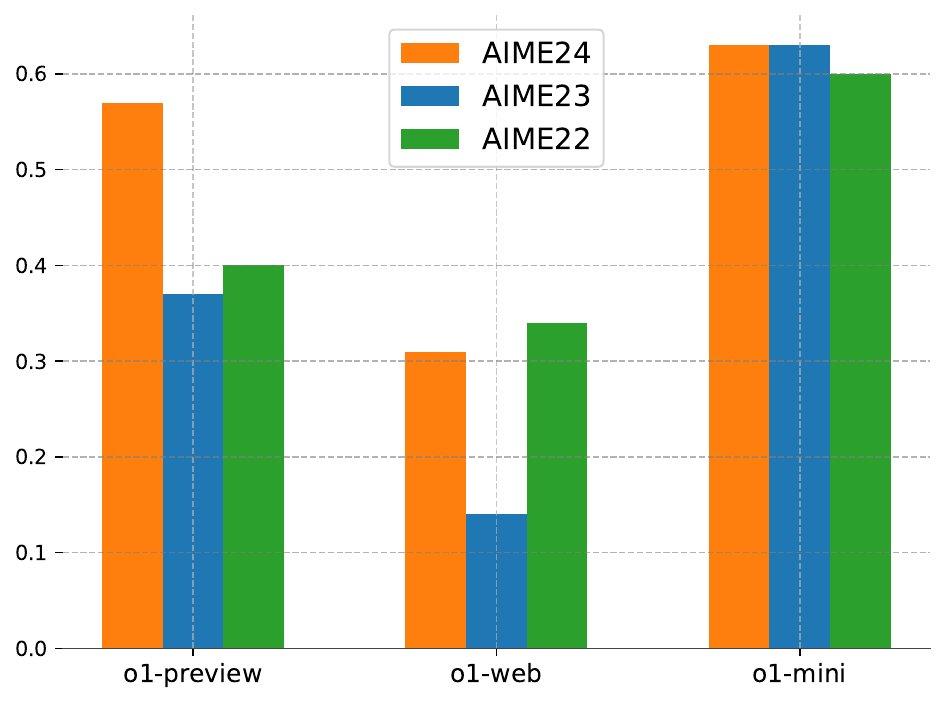}
    \caption{The results of o1 model on AIME24, AIME23, and AIME22. }
    \label{fig:AIME_result}
\end{wrapfigure}

\subsection{The Search Space also Determines the Upper Boundary of LLMs}
Apart from the Agent Workflow, BoN also performs relatively well across various datasets, but its performance is limited by $N$. 
To fully explore the upper bound of BoN's capabilities, we increased the value of N in HotpotQA.
For the sake of comprehensively evaluating the BoN's capability based on different LLMs with different capability levels, we also evaluate Qwen2.5-72B and Llama3-70B in Fig.~\ref{fig:BoN}.
Specifically,
as shown in Fig.~\ref{fig:BoN}, we compare the results of BoN using different backbone models under different search spaces (i.e., $N=1,4,8,16$).
With the $N$ increasing, the performance of BoN tends to stabilize. Notably, both Qwen2.5 and Llama3 achieve excellent performance on the HotpotQA dataset. However, when BoN uses these three models as backbone models, performance does not improve consistently with increasing $N$. When $N > 8$, the performance of the models either stabilizes or declines,
and we suppose the reason is that the performance of the search methods is jointly related to the reward model and searching space.

\begin{figure}[t]
    \centering
    \includegraphics[width=0.8\linewidth]{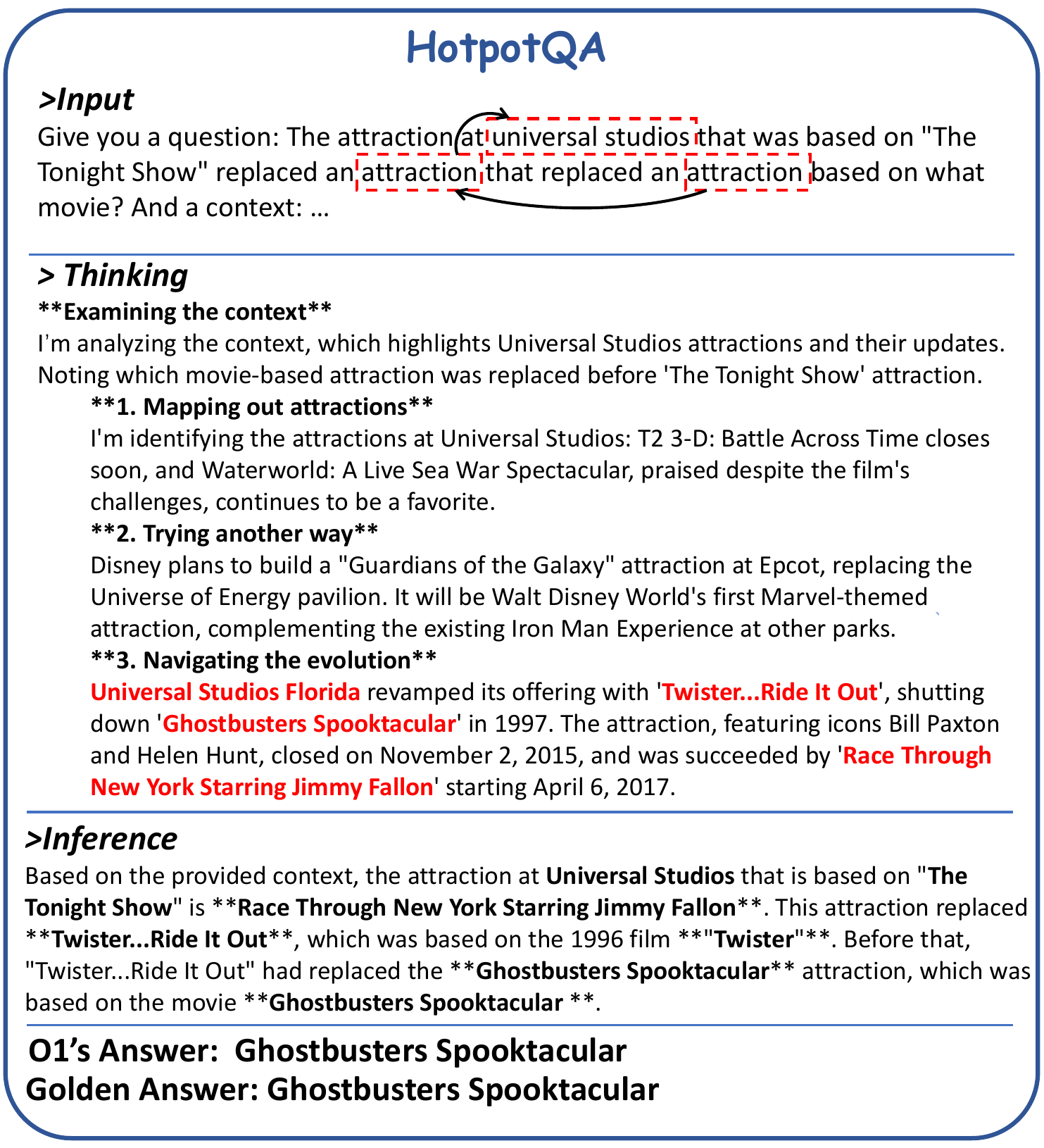}
    \caption{The o1's case of HotpotQA. }
    \label{fig:case_hotpotqa}
\end{figure}

\subsection{Analysis on Data Filter}
\label{App: Analysis on Data Filter}
\begin{figure}[t]
    \centering
    \includegraphics[width=0.99\linewidth]{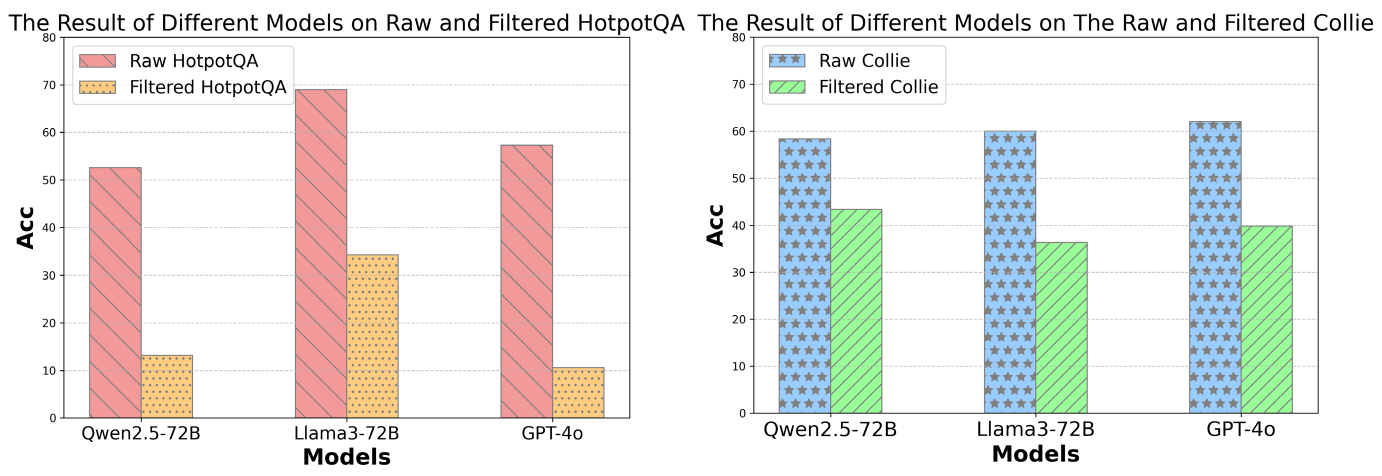}
    \caption{The results of the LLMs on the raw bench and the filtered bench. On the left subfigure, we present LLMs' capabilities on the raw and filtered HotpotQA, and on the right subfigure, we provide the corresponding results on Collie. }
    \label{fig:data_filter}
\end{figure}

\begin{wraptable}{r}{0.47\textwidth} 
\centering
\small
\begin{tabular}{cc|c|c}
\toprule
 \multicolumn{2}{c|}{\textbf{Commonsense Reasoning}} & \multicolumn{1}{c|}{\textbf{Coding}} & \multicolumn{1}{c}{\textbf{Math}}  \\
 HotpotQA & Collie  & USACO & AIME \\
\midrule
\midrule
274& 226 & 139 & 90 \\
\bottomrule
\end{tabular}
\caption{The statistics of filtered benchmarks.}
\label{table:static}
\end{wraptable}

The current benchmarks contain many simple samples that cannot distinguish the performance difference across different LLMs, and we filter the samples in HotpotQA and Collie.
To demonstrate the impact of our data filter module, we compare the performance of LLMs on benchmarks before and after applying the data filter,
where Table~\ref{table:static} presents the statistics about our selected benchmarks after data filtering.
As shown in Fig.~\ref{fig:data_filter}, after data filtering, the scores of different LLMs are relatively lower and show greater distinction. Notably, on HotpotQA, the differences between Qwen2.5 and GPT-4o become evident on our filtered benchmark,
which demonstrates the effect of our data filter strategy.

\subsection{The Math Ability of OpenAI's o1}

To comprehensively evaluate the o1 models' ability in math, we evaluate the o1-preview, o1-web, and o1-mini on the AIME22, AIME23, and AIME24.
It is worth mentioning that the o1-mini demonstrates the best performance (around 60\%) across these three datasets.
However, the performance of the o1-preview fluctuates significantly across different datasets.
For example, for o1-preview, the best performance on AIME24 is 57\%, while results on the other two datasets are both around 40\%.

\section{Case Study}

\begin{figure}[t]
    \centering
    \includegraphics[width=0.8\linewidth]{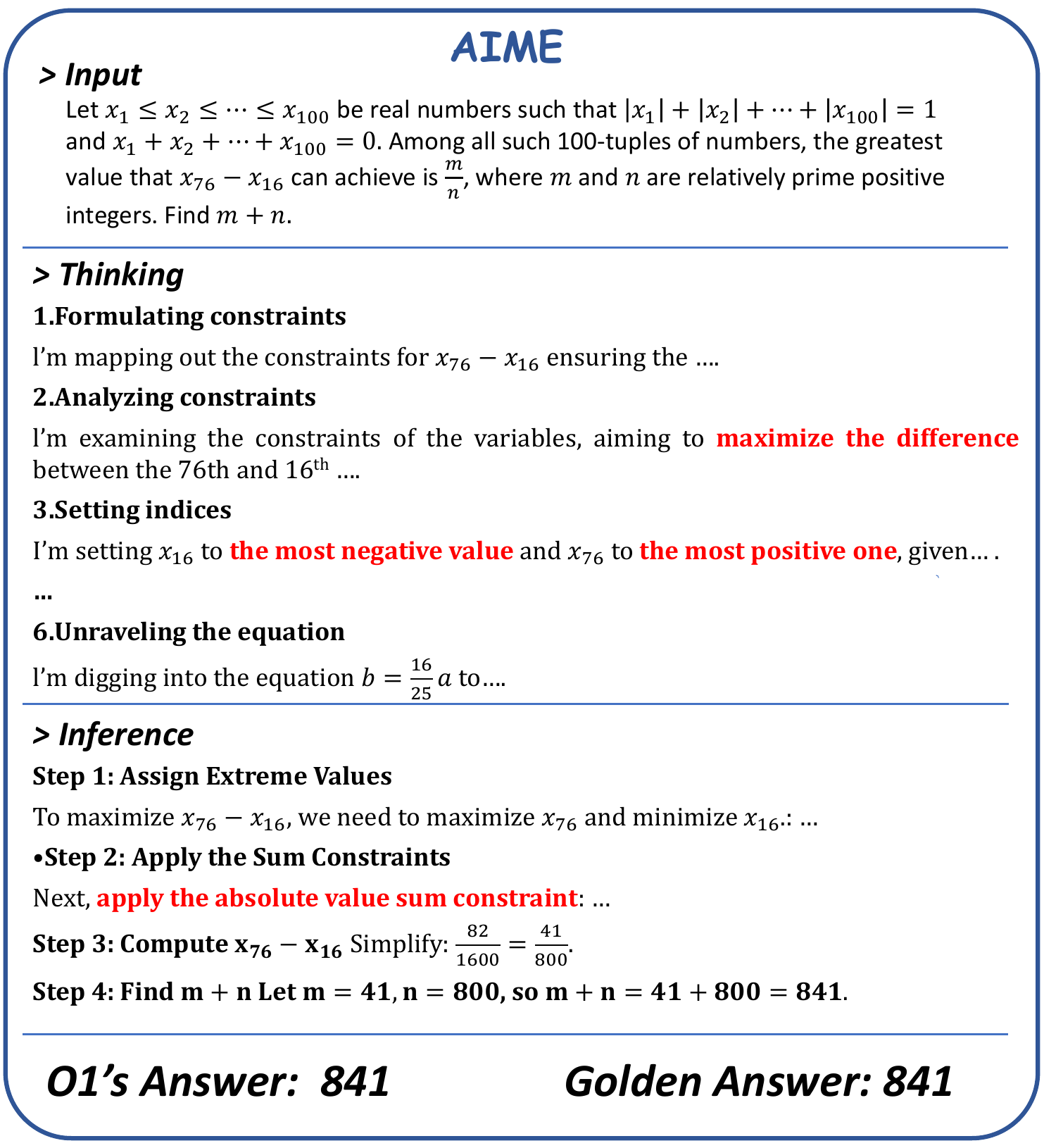}
    \caption{The o1's case of AIME. }
    \label{fig:case_aime}
\end{figure}

The HotpotQA needs LLMs to process multiple documents to obtain the results of the reasoning question, which needs multi-hop reasoning.
The o1 generally summarizes the content of the documents from different perspectives to arrive at a solution.
As shown in Fig.~\ref{fig:case_hotpotqa}, firstly, the o1 model proposes the ``main idea'' (i.e., analyzing the context related to the question).
Then it obtains the content in the three dimensions (i.e., ``Mapping out attractions'', ``Mapping out attractions'', and ``Navigating the evolution''). It is worth mentioning that ``Navigating the evolution'' contains the reasoning chain of the multi-hot reasoning question.

\subsection{AIME}
Unlike simpler methods that tackle problems through straightforward subtasks, AIME demands a deep integration of diverse mathematical principles to derive accurate solutions. As illustrated in Fig.~\ref{fig:case_aime}, o1 uses a CoT method following a structured approach: ``Identify Key Concepts'', ``Analyze Constraints'', ``Apply Mathematical Formulas'', and  ``Construct Logical Reasoning'',
where the ``Logical Reasoning'' involves a systematic and detailed process for solving multi-step problems. This reasoning pattern emphasizes the importance of integrating insights from various mathematical concepts to enhance the model's ability to handle complex problems effectively. By following this framework, LLMs can navigate the intricate nature of AIME problems, ensuring that all relevant mathematical concepts and logical steps are thoroughly considered. 

\subsection{Collie}
\begin{figure}[t]
    \centering
    \includegraphics[width=0.8\linewidth]{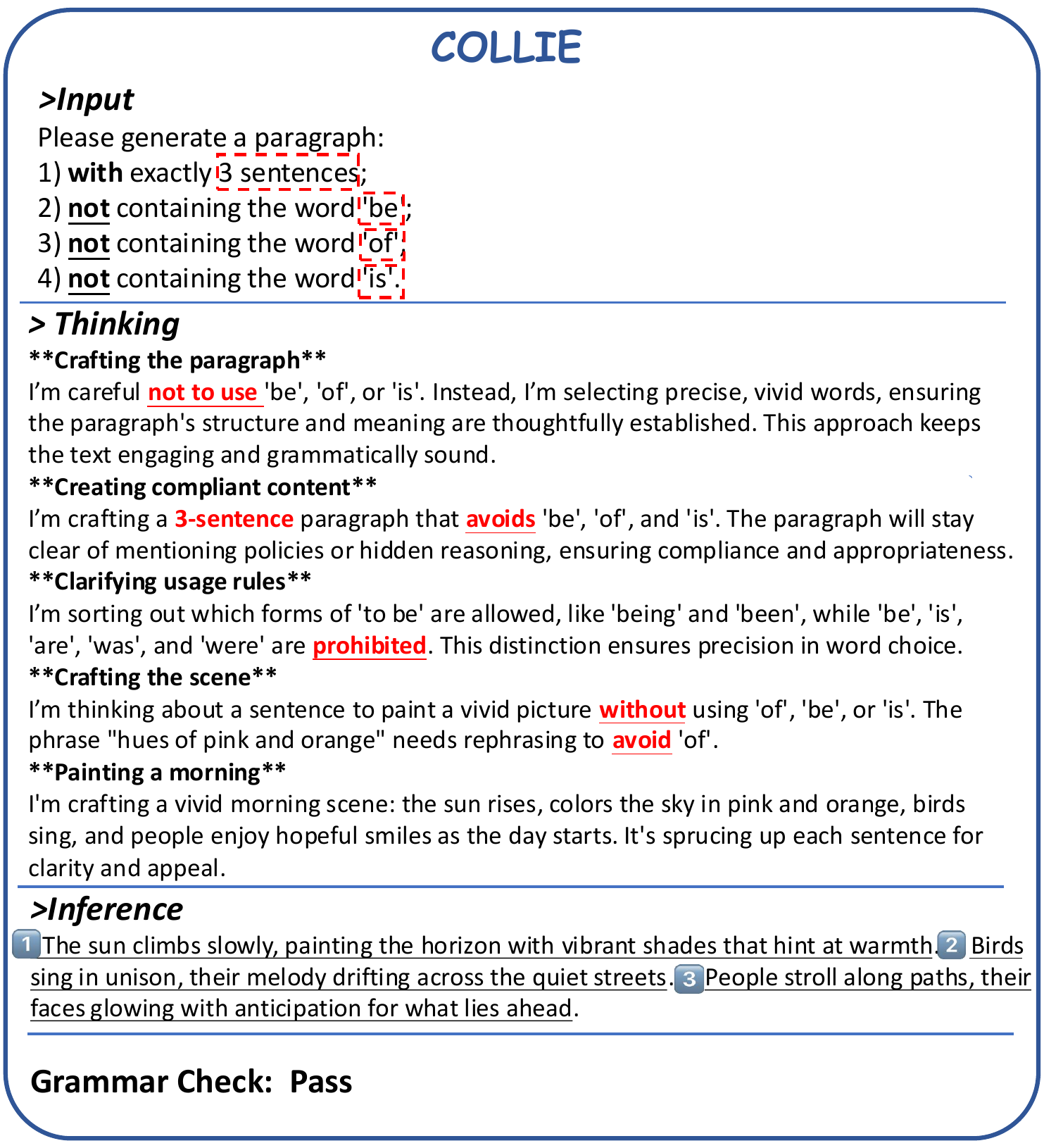}
    \caption{The o1's case of Collie. }
    \label{fig:case_collie}
\end{figure}

The task of the Collie dataset requires generating a paragraph based on the constraint.
Specifically,
as illustrated in Fig.~\ref{fig:case_collie}, at each step of the reasoning process of the o1 model, the form of the constraint is emphasized to strengthen for better generation.
Therefore,
for tasks requiring strictly controlled generation, the model needs to emphasize the instructions multiple times to guide the content generation.

\subsection{USACO}
\begin{figure}[t]
    \centering
    \includegraphics[width=0.8\linewidth]{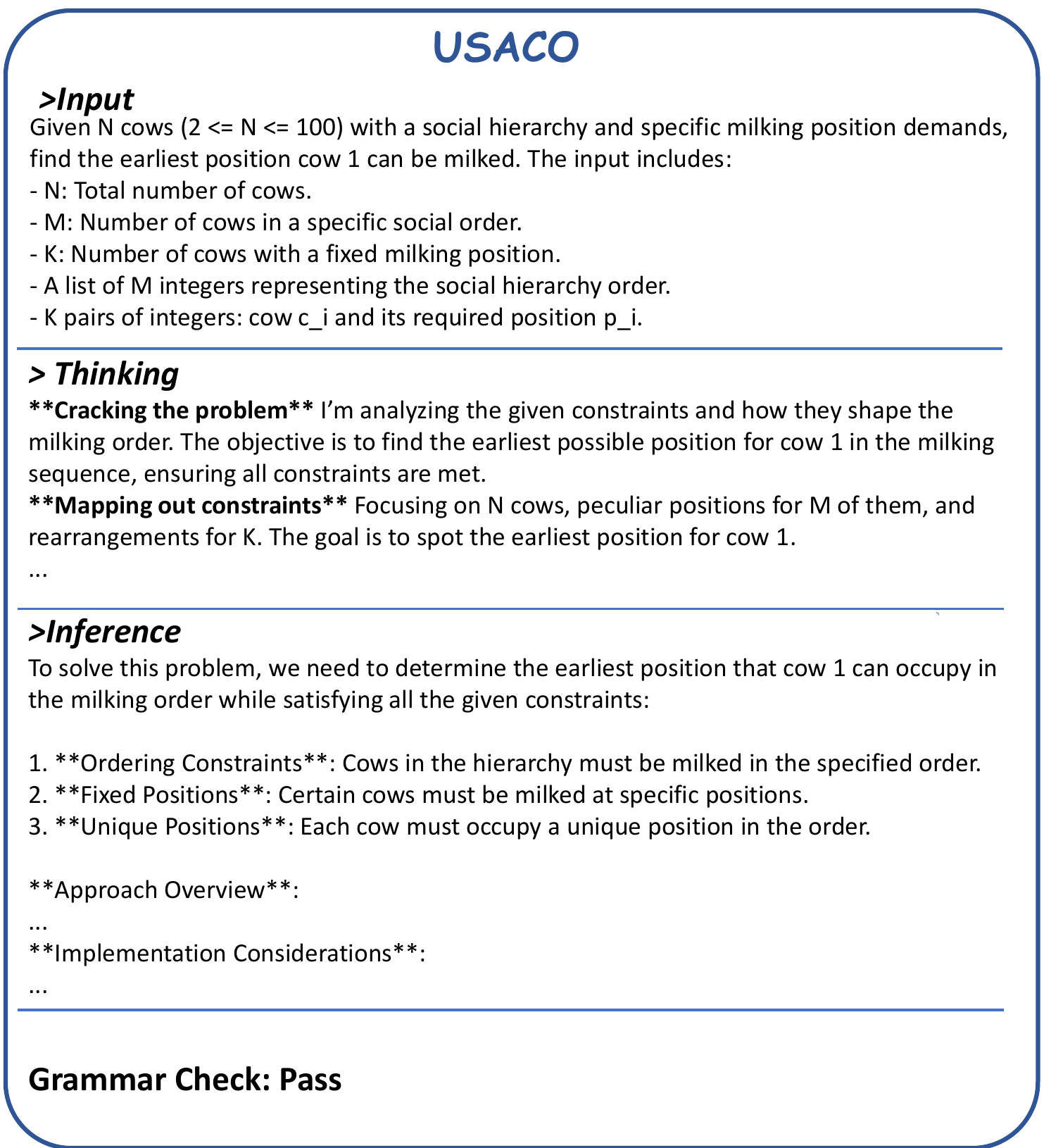}
    \caption{The o1's case of USACO.}
    \label{fig:usaco_example}
\end{figure}

In the USACO competition, o1 demonstrates its robust problem-solving capabilities. As illustrated in Fig.~\ref{fig:usaco_example}, o1 starts by establishing a foundational framework by defining key variables and data structures, and then applies algorithmic logic for state transitions, which will produce the optimal solution gradually. 
Besides, o1 not only considers all possible paths and scenarios but also uses loops, recursion, and other methods to verify each step rigorously, which helps o1 comprehensively cover multiple aspects of the problems and generate correct solutions efficiently. 

\section{Conclusion}
In this work, we explore the capabilities of OpenAI's o1 model in tasks involving mathematics, coding, and commonsense reasoning, and compare o1 with previous Test-time Compute methods (BoN, Step-wise BoN, Self-Refine, and Agent Workflow),
where the findings are as follows.
First, we find that o1 achieves better results than other Test-time Compute methods across most tasks.
Second, the Agent Workflow method shows significant improvements in all tasks, but the BoN, Step-wise BoN, and Self-Refine methods yield limited improvements due to their reliance on the model's long-context instruction-following ability and the performance of the reward model.
Third, we also summarize six reasoning patterns (i.e., SA, MR, DC, SR, CI, EC) of o1, and demonstrate that SR and  DC are the main reasoning patterns of o1, which indicates that these reasoning patterns are crucial for improving reasoning performance.
Finally, we hope our study on the reasoning patterns of o1 can guide developers and researchers in understanding the principle of o1 and facilitate the growth of foundation models.

\clearpage
\bibliography{iclr2025_conference}
\bibliographystyle{iclr2025_conference}

\end{document}